\documentclass[runningheads]{llncs}

\usepackage[utf8]{inputenc}                     
\usepackage[numbers,sectionbib]{natbib}         
\usepackage{graphicx}                           
\usepackage{amsmath,amssymb}                    
\usepackage{booktabs}                           
\usepackage{import}                             
\usepackage{xcolor}                             
\usepackage{siunitx}                            
\usepackage{subcaption}                         
\usepackage{pdflscape}                          
\usepackage{wrapfig}                            
\usepackage{tikz}                               
\usepackage{pgfplots}                           
\pgfplotsset{compat=1.14}                       
\usetikzlibrary{calc,fit,positioning,chains,patterns,decorations.pathreplacing,decorations.text,shapes,backgrounds,external,spy,arrows.meta}
\usepgfplotslibrary{external,patchplots}
\usepackage{hyperref}                           
\usepackage[capitalise,noabbrev]{cleveref}      
\usepackage{csquotes}                           

\begin{document}

\title{Visual Mesh: Real-time Object Detection Using Constant Sample Density}

\titlerunning{Visual Mesh}
\authorrunning{T. Houliston \and S. K. Chalup}

\author{
	Trent Houliston \orcidID{0000-0002-7744-0472}
	\and \\ Stephan K. Chalup \orcidID{0000-0002-7886-3653}
}

\institute{School of Electrical Engineering and Computing \\
The University of Newcastle, Callaghan, NSW, 2308, Australia. \\
\texttt{\url{trent@houliston.me}}, \texttt{\url{stephan.chalup@newcastle.edu.au}}}

\date{\today}
\maketitle              
\begin{abstract}
This paper proposes an enhancement of convolutional neural networks for object detection in resource-constrained robotics through a geometric input transformation called Visual Mesh.
It uses object geometry to create a graph in vision space, reducing computational complexity by normalizing the pixel and feature density of objects.
The experiments compare the Visual Mesh with several other fast convolutional neural networks.
The results demonstrate execution times sixteen times quicker than the fastest competitor tested, while achieving outstanding accuracy.

\keywords{Convolutional Neural Network \and Deep Learning \and Ball Detection \and Graph Transformation \and TensorFlow \and Machine Vision}
\end{abstract}

\section{Introduction}

	This paper introduces a Visual Mesh that defines an input transformation for convolutional neural networks (CNN).
	By normalizing object size, the Visual Mesh accounts for differences in an object's appearance when detecting and localizing it.
	This allows simpler network architectures to be used and reduces oversampling, improving the computational performance substantially.

	CNNs require powerful hardware to perform in real-time.
	Despite this, some networks have been developed to run on constrained hardware with limited success.
	\citet{speck_ball_2016} built a CNN for detecting the coordinates of a soccer ball on an image.
	When implemented on their target platform it ran in \SI{26}{\milli\second} and had an accuracy of \SI{58}{\percent} in $x$ and \SI{52}{\percent} in $y$.
	The accuracy dropped to less than \SI{30}{\percent} in distances over two meters.
	Therefore, this approach had limited success in object localization.

	Faster and more accurate systems have been developed that only perform object classification.
	These systems utilize color segmentation to provide proposals for a CNN to classify.
	As a result they were much faster than systems that localize objects, however, color segmentation is sensitive to changes in lighting conditions and must be manually calibrated.
	\citet{javadi_humanoid_2017} utilized such a system for detecting humanoid robots.
	The best performing network ran in \SI{2.36}{\milli\second} with \SI{97.56}{\percent} accuracy per proposal on an Intel Core i5 \SI{2.5}{\giga\hertz}.
	\citet{cruz_using_2017} developed a system to classify Aldebaran NAO robots.
	This network executed in $\approx$\SI{1}{\milli\second} per proposal.
	\citet{albani_a-deep_2017} and \citet{bloisi_machine_2017} utilized a similar technique for ball detection.
	This system was implemented on an Aldebaran NAO robot and processed 14-22 frames per second as the only process running.
	The reliance on color segmentation for proposals limits these networks to color coded environments.

	\begin{figure}[htbp]
		\centering
		\begin{subfigure}[t]{0.34\textwidth}
			\centering
			\includegraphics[width=\linewidth]{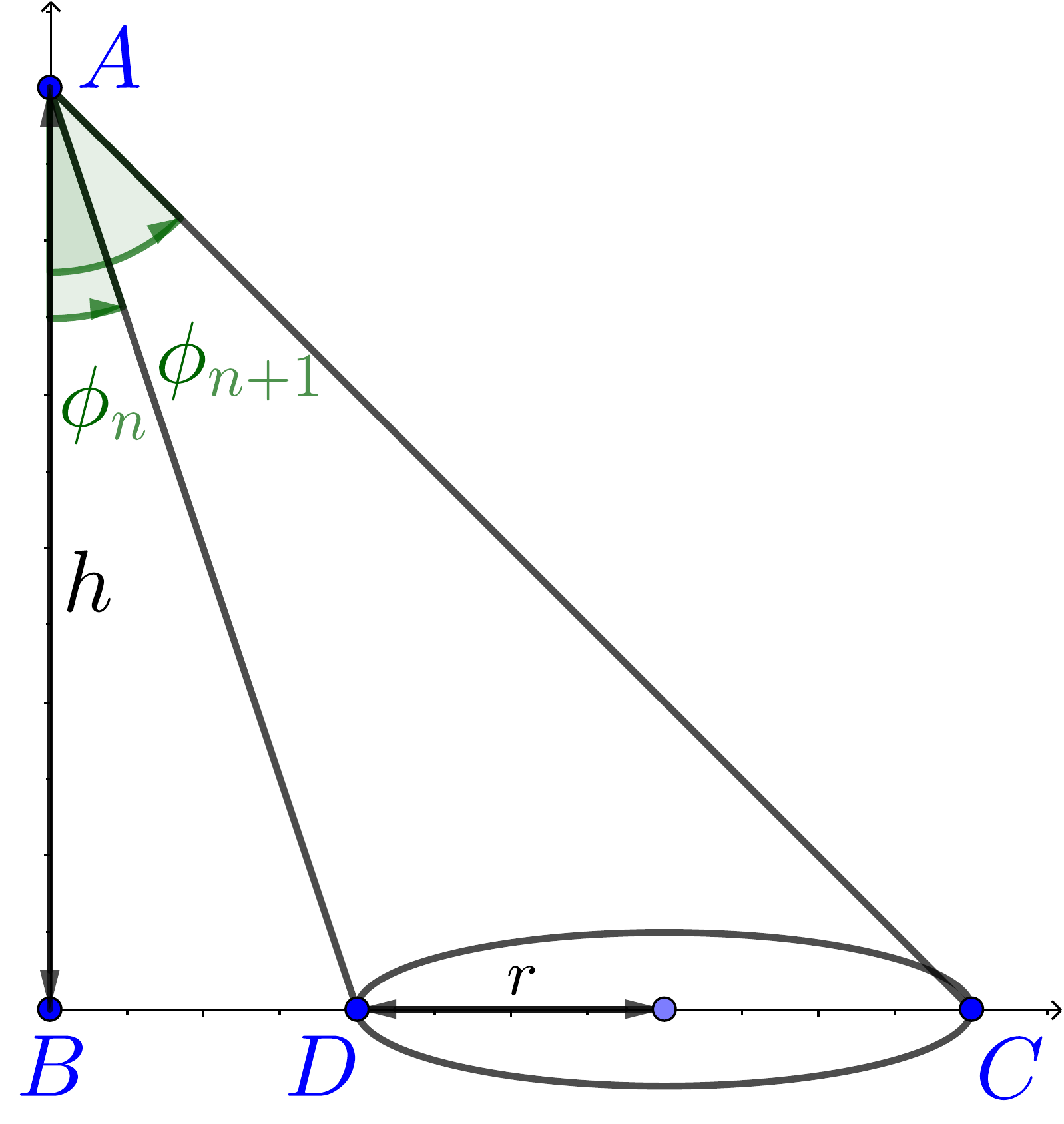}
			\caption{
				\label{fig:circle_phi}
				Side view: $\phi_{n+1}$ for a circle on the observation plane is found using \cref{eqn:circle_phi1}.
			}
		\end{subfigure}
		\hfill
		\begin{subfigure}[t]{0.37\textwidth}
			\centering
			\includegraphics[width=\linewidth]{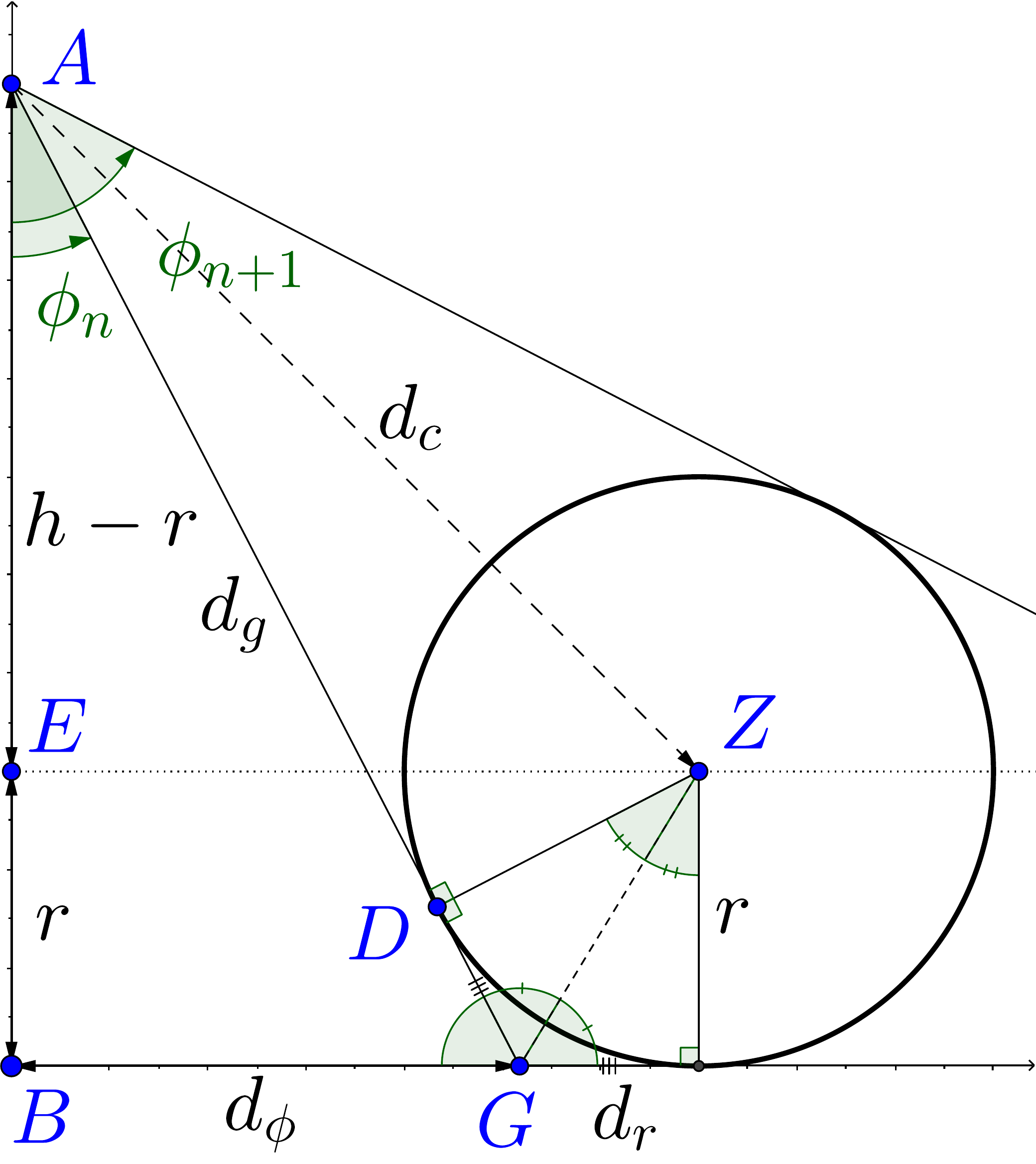}
			\caption{
				\label{fig:sphere_phi}
				$\phi_{n+1}$ for a sphere can be found by inspecting its tangent lines and using \cref{eqn:sphere_phi_1}.
			}
		\end{subfigure}
		\hfill
		\begin{subfigure}[t]{0.22\textwidth}
			\centering
			\includegraphics[width=\linewidth]{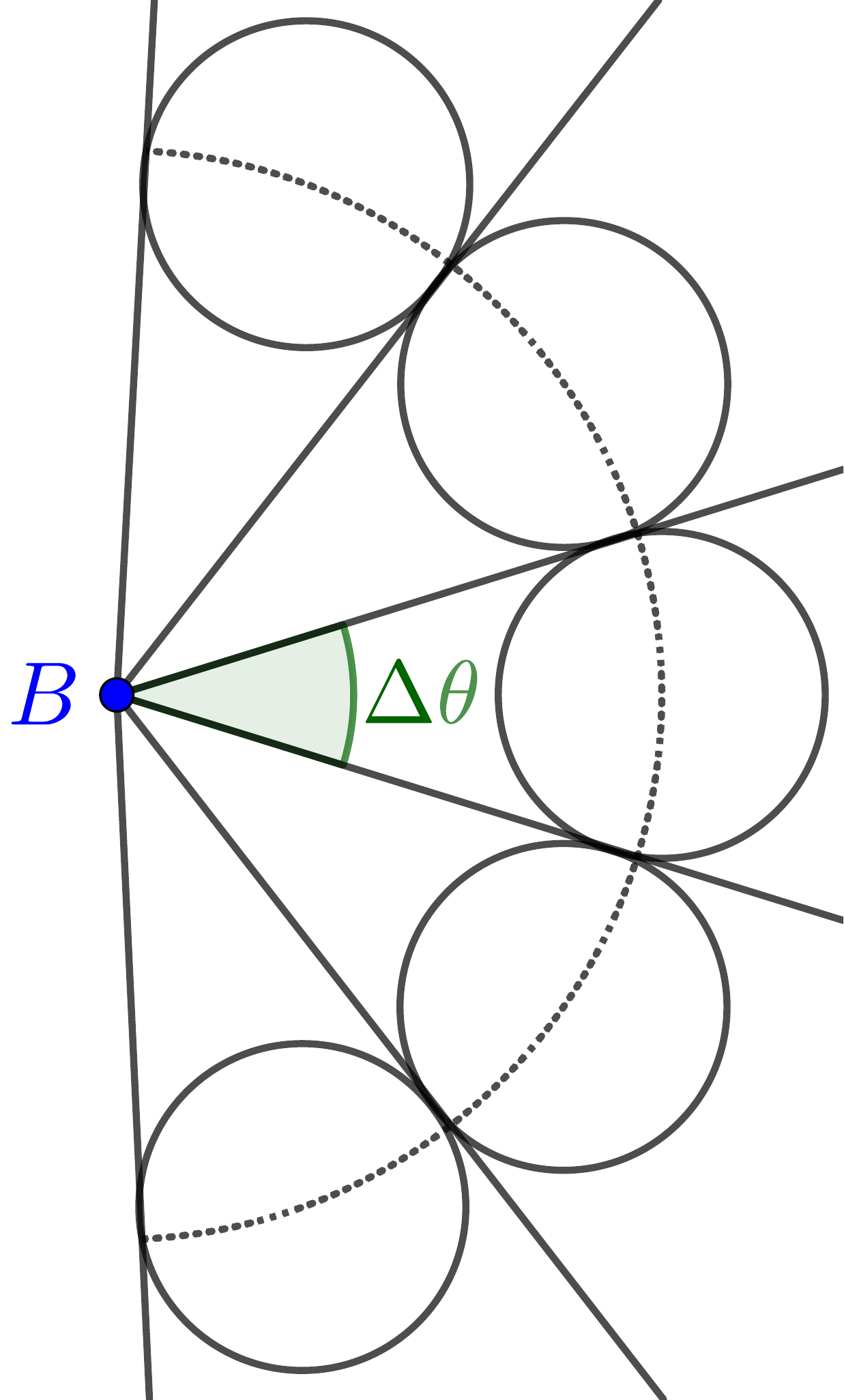}
			\caption{
				\label{fig:circle_theta}
				$\Delta\theta_n$ segments in top view of the observation plane.
			}
		\end{subfigure}
		\caption{
			\label{fig:geometry}
			Geometry for calculating $\phi_{n+1}$ and $\Delta\theta_n$.
		}
	\end{figure}

\section{Visual Mesh Geometry}
	The Visual Mesh detects objects that lie on a plane at a known distance and orientation from the camera.
	This plane is referred to as the observation plane.
	The geometry of the Visual Mesh can be described using \cref{fig:geometry} where the camera is assumed to be at point $A$.
	The target object's geometry determines the pixel resolution, i.e., the placement of points in the Visual Mesh.
	The geometry for two target object shapes are analyzed in this paper:
	Circles are appropriate for detecting two-dimensional objects on the observation plane (\cref{fig:circle_phi}).
	Spheres are appropriate for three-dimensional objects that have an approximately equal extension in all dimensions (\cref{fig:sphere_phi}).
	More complex objects such as cylinders could also be modeled.

	For establishing the Visual Mesh two orthogonal angular components have to be determined.
	These are $\Delta\phi_n$ and $\Delta\theta_n$ and are given by the angular diameters of the target object with respect to points $A$ and $B$.
	The height $h$ of the camera above the observation plane and the radius $r$ are required to calculate the mesh.

	The first component is $\Delta\phi_n:= \phi_{n+1} -\phi_n$ and is determined by the inclinations $\phi_n$ from directly below the camera.
	A series $\phi_n$, $n = 0,..,N$ is given recursively by function $f:\mathbb{R}\rightarrow\mathbb{R}$, $\phi_{n+1} = f(\phi_{n}) = \phi_{n} + \Delta\phi_n$  where $\phi_0 = 0$.

	The second component, $\Delta\theta_n$, is measured around point $B$ in the observation plane and depends on $\phi_n$ for both, circle and sphere objects (\cref{fig:circle_theta}).

	The inclinations $(\phi_n)_{n=0,...N}$ induce a series of nested concentric cones with vertex at $A$ and center axis orthogonal to the observation plane.
	Each of these cones is radially segmented at its basis by~$\Delta\theta_n$ and the tangent rays from~$B$.

	\subsection{Circle}
		The geometry for circles is shown in \cref{fig:circle_phi}.
		$\phi_{n+1}$ for a circle is calculated by adding the diameter $2r$ of the circle to its distance $\overline{BD}$ to obtain
		\begin{equation}
			\label{eqn:circle_phi1}
			\phi_{n+1} = \tan^{-1}\left(\tan\left(\phi_n\right) + \frac{2r}{h}\right)
		\end{equation}
		\cref{fig:circle_theta} shows the geometry for $\Delta\theta_n$ within the 2D observation plane where
		\begin{equation}
			\label{eqn:circle_dtheta}
			\Delta\theta_n = 2\sin^{-1}\left(\frac{r}{h\tan\left(\phi_n\right)}\right)
		\end{equation}
		This formulation of $\Delta\theta_n$ has a singularity when the center of the object is closer than its radius making it more difficult for the mesh to detect objects directly below the camera.

	\subsection{Sphere}
		For spheres $\Delta\phi_n$ is determined by the sphere's shadow from a virtual light at $A$ and it decreases more slowly with $n$ than for circles. \cref{fig:sphere_phi} shows how $\phi_{n+1}$ is calculated.
		Using the triangle $\triangle AEZ$ and edges $\overline{AE}$ and $\overline{EZ}$ gives
		\begin{align}
			\begin{aligned}
				\label{eqn:sphere_phi_1}
				\phi_{n+1} & = 2\tan^{-1}\left(\frac{d_{\phi} + d_{r}}{h - r}\right) - \phi_n                                  \\
				           & = 2\tan^{-1}\left(\frac{r\sec \left( \phi_n \right)}{h-r}+\tan \left( \phi_n \right) \right) - \phi_n
			\end{aligned}
		\end{align}
	The calculation of $\Delta\theta_n$ is the same as for circles and uses \cref{eqn:circle_dtheta}.

	\subsection{Object Dependent Sample Density}
		The current description guarantees one point in the mesh for the target object.
		For use in computer vision, multiple sample points per object are required.
		Let's assume our object requires $k$ pixels to be recognizable.
		In the Visual Mesh this $k$ corresponds to the number of intersections of the $\phi_n$ rings with the object.
		A~$\phi_n$ ring is obtained by rotating vector $\overrightarrow{AD}$ about the axis $\overrightarrow{AB}$.
		If $\Delta\phi_n$ and $\Delta\theta_n$ are reduced, the spacing between the $\phi_n$ rings will be decreased which leads to more intersections with the target object.

		\begin{wrapfigure}[16]{r}{0.37\textwidth}
			\centering
			\includegraphics[width=\linewidth]{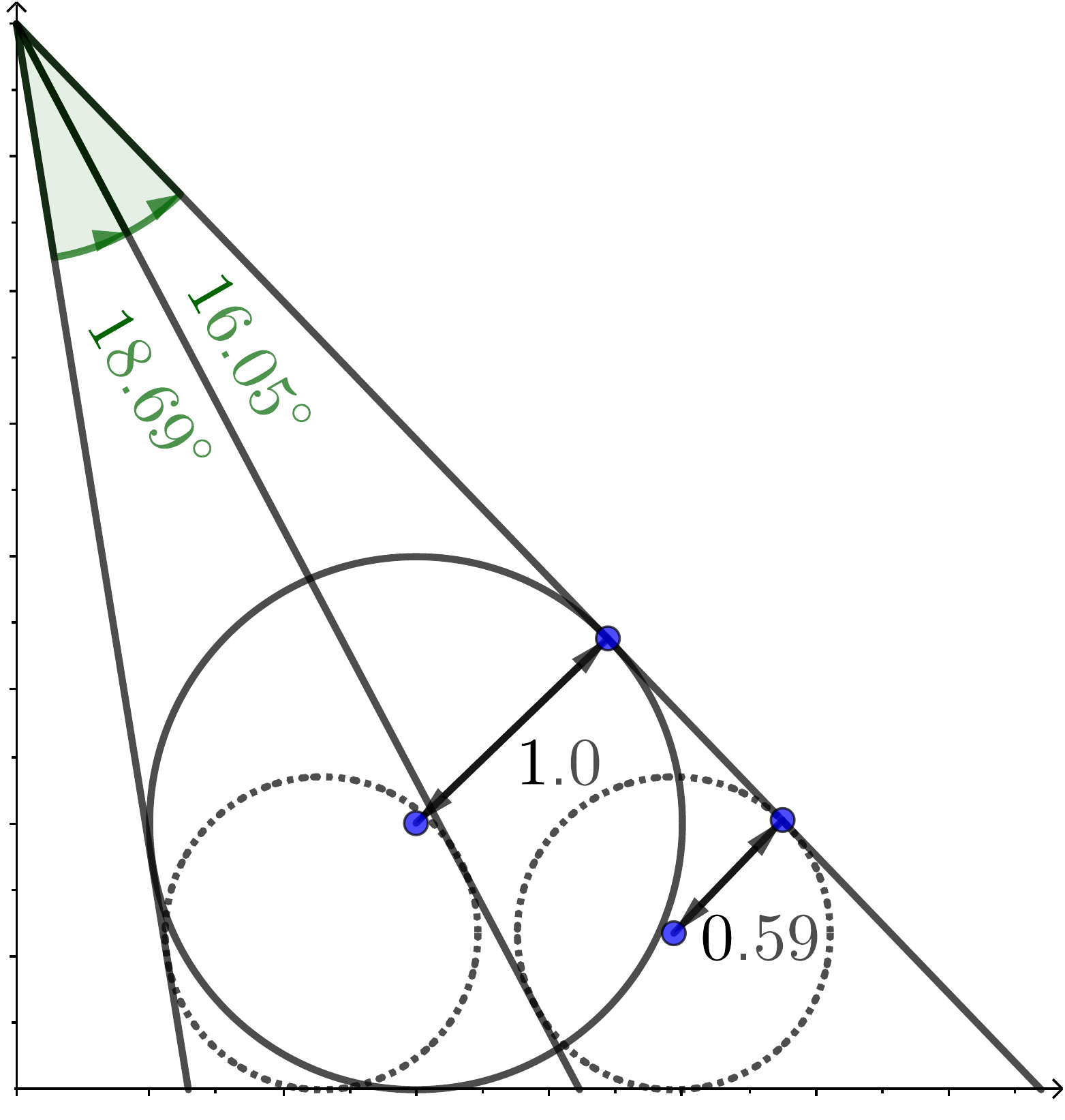}
			\caption{
				\label{fig:multi_sphere_intersections}
				Multiple sample points on a sphere can be calculated by finding the smaller sphere's radius.
			}
		\end{wrapfigure}

		An increase in the number of sample points for the circle model can be achieved by dividing $\Delta\theta_n$ in (\ref{eqn:circle_dtheta}) by $k$ and also the diameter of the circle by $k$, i.e., replacing $2r$ in (\ref{eqn:circle_phi1}) by $2r/k$.

		In the sphere model $k$ sample points can be achieved by creating a version of the mesh where the original sphere is replaced by smaller spheres so that the original sphere intersects with $k$ $\phi_n$ rings associated with the smaller spheres (\cref{fig:multi_sphere_intersections}).
		If $k$ is expressed as fraction $k = \frac{p}{q}$, $p,q\in\mathbb{N}-\{0\}$, the equation relating the radii of the spheres is given by $f^{q}\left(\phi_0, r_0\right) = f^{pq}\left(\phi_0, r_1\right)$ where $r_0$ is the radius of the target and $r_1$ is the radius of the small spheres in the mesh. A solution for $r_1$ can be obtained numerically.

	\subsection{Graph Structure of the Mesh}
		\begin{wrapfigure}[18]{l}[0pt]{0.5\textwidth}
			\centering
			\subimport{figures/project_mesh/}{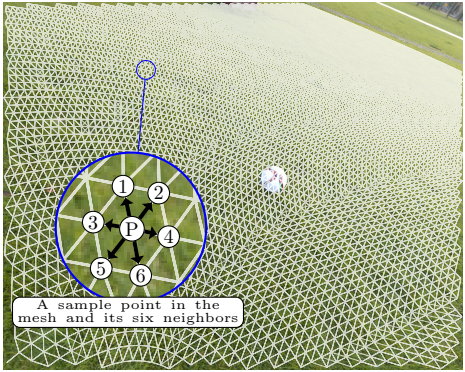}
			\caption{
				\label{fig:mesh_projection}
				The Visual Mesh projected onto an image.
				Note that four $\phi_n$ rings pass through the ball regardless of its location.
			}
		\end{wrapfigure}

		A mesh can be generated using the points around $\phi_n$ rings (see arcs in \cref{fig:mesh_projection}).
		In each $\phi_n$ ring, points are separated by $\Delta\theta_n$.
		This ensures that the number of points within an object falls within a small range ($\pm 1$ in $\phi$ and $\theta$).
		Each point is connected with edges to the two adjacent points on the same $\phi_n$ ring as well as to the two nearest points on the $\phi_{n\pm1}$ rings.
		The single point below the camera is connected by six equally spaced points.
		Projecting these points onto an image creates a mesh structure as shown in \cref{fig:mesh_projection}.

		Another method to view the mesh is to project the $\phi_n$ rings onto concentric circles as in \cref{fig:mesh_distance}.
		Due to perspective, the size of objects decreases in distance within the original image, while the Visual Mesh ensures objects are always the same size.

		\begin{figure}[htbp]
			\centering
			\includegraphics[width=0.45\textwidth]{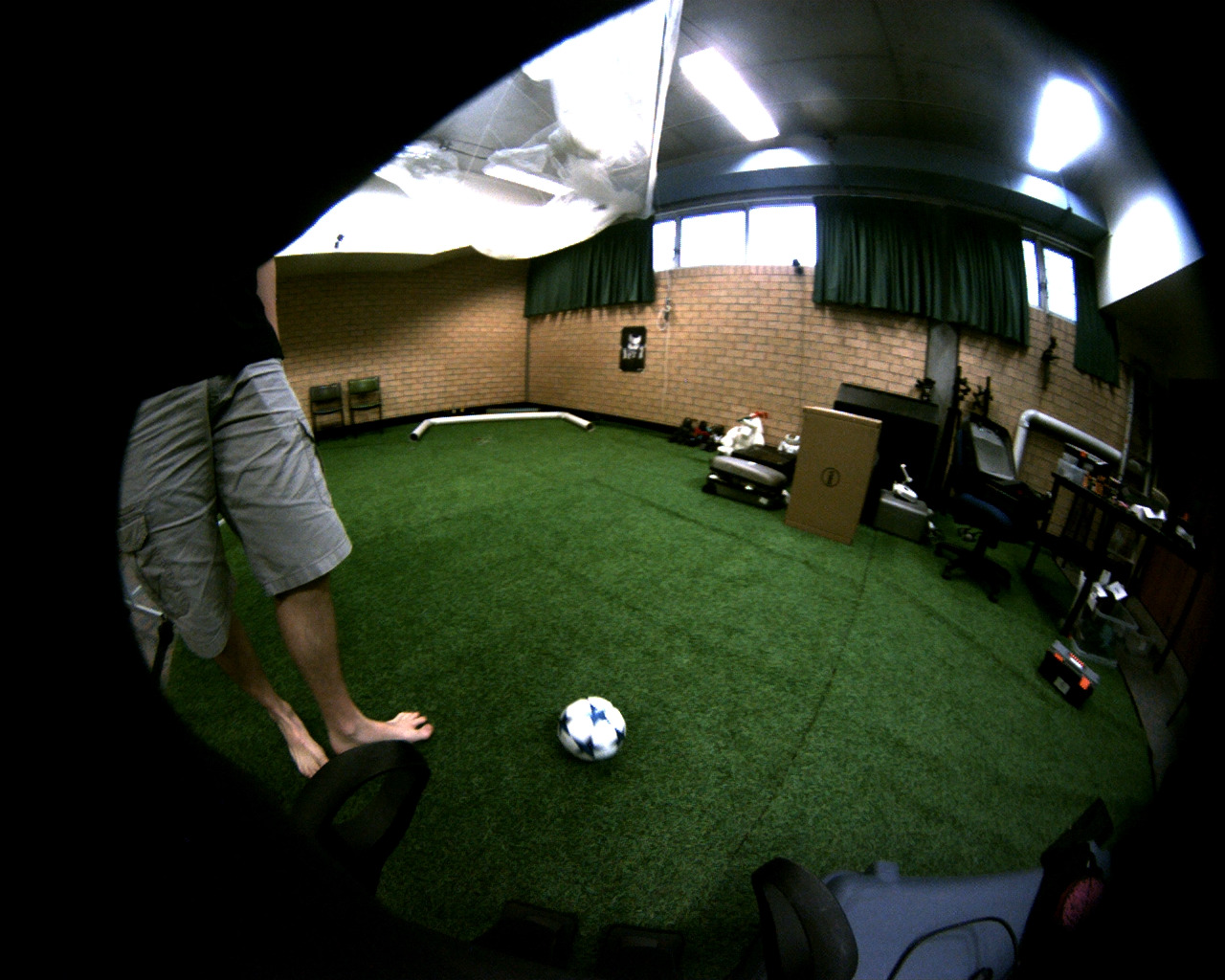}
			\includegraphics[width=0.45\textwidth]{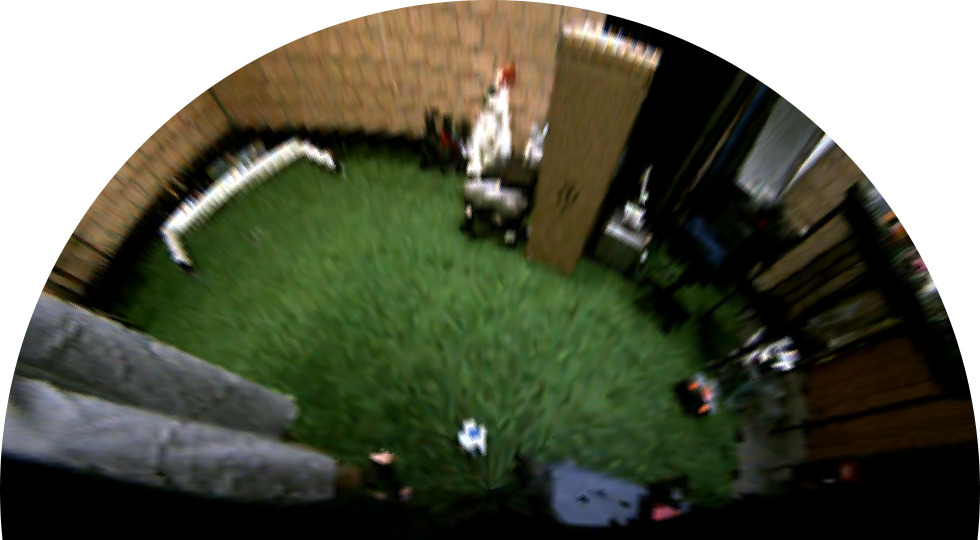} \\
			\includegraphics[width=0.45\textwidth]{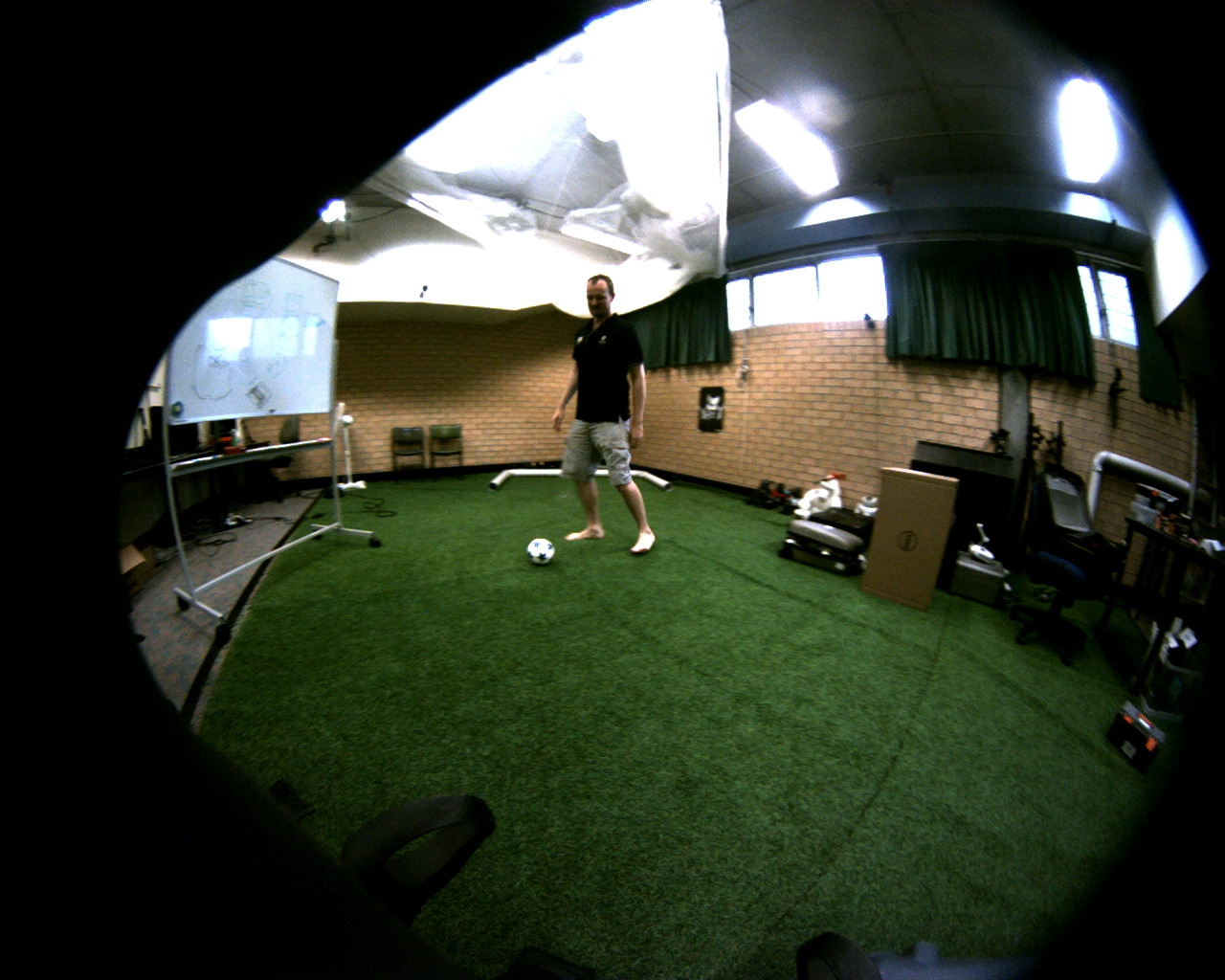}
			\includegraphics[width=0.45\textwidth]{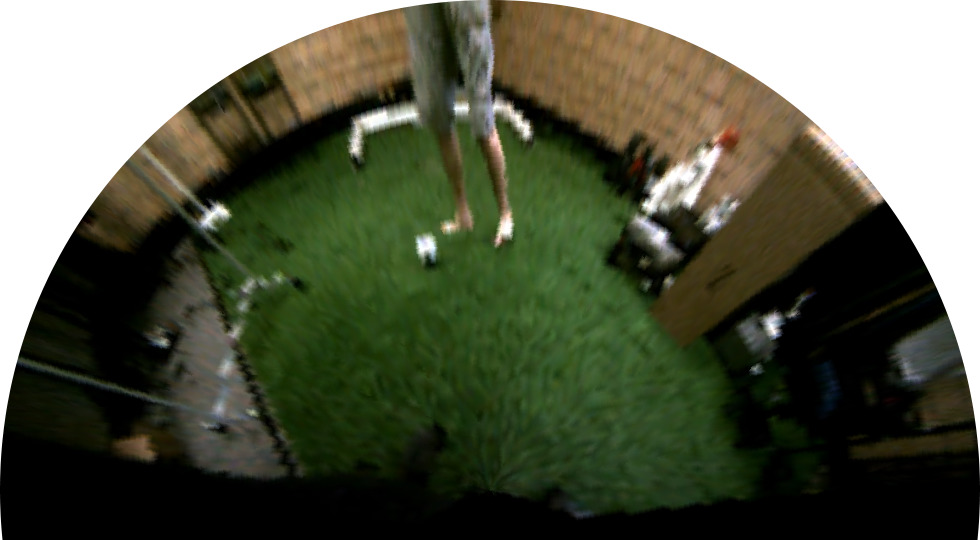} \\
			\caption{
				\label{fig:mesh_distance}
				The Visual Mesh projected in concentric rings.
				Due to perspective, the size of the ball decreases with distance in the original (left).
				In the Visual Mesh, the ball has always a similar size (right).
			}
		\end{figure}

	\subsection{Network}
		Once the image data has been transformed by the Visual Mesh, it exists as a graph, rather than a grid of pixels.
		The pixels no longer have nine neighbors, but six.
		This changes how convolutions occur when executed on the graph.

		For example, a $3\times 3$ convolution in a typical CNN accesses eight pixels around a central pixel.
		The equivalent operation in the graph accesses points with a graph distance of one, its six neighbors.
		This has a positive impact on performance, as two fewer values need to be considered.
		Larger convolutions would be equivalent to operating on points at a larger maximum graph distance.
		For example, a $5\times 5$ convolution would operate on all points that have a maximum graph distance of two to the central point.

\section{Evaluation of the Visual Mesh}

	\subsection{Dataset}
		A semi-synthetic dataset with masks that segment out the ball was created for training.
		By using \SI{360}{\degree} high dynamic range (HDR) images to provide image-based lighting, along with physics-based rendering, realistic semi-synthetic scenes were generated.
		From this, the mask images, as well as the camera orientation and position can be obtained.

		Using a number of different HDR scenes taken from RoboCup 2017, the NUbots' laboratory and online\footnote{HDRI Haven \url{https://hdrihaven.com/}}, as well as over a hundred different soccer ball designs, over 160,000 images were generated.
		These soccer ball designs were not limited to 50\% white as per RoboCup rules and included balls of various colors.
		Each of these images varied the position of the soccer ball and switched between equisolid and rectilinear camera projections.

		The distance of the balls from the camera varied between zero and ten meters.
		The intensity of the lighting varied in the scene.
		The rendered soccer ball was selected from a set of 140 different models.
		The distribution of distances was designed to provide a uniform variation in the pixel size of the ball.
		This allowed a consistent variation in the angular diameter of the ball in the image.
		It prevented a large number of visually small balls that would have occurred with a uniform distribution over distance.

		\subsection{Network Architecture}
			Each node in the Visual Mesh performed a $3\times 3$ convolution using its six neighbors.
			These layers were stacked to varying depths from two to nine and with output widths varying from two to eight, resulting in a fully convolutional net.

			Networks of width four performed significantly better than networks of other widths as the hardware utilized can vectorize on four elements.
			The results discussed in \cref{sec:visual_mesh_results} only include network widths of four.

			Networks were also tested with ReLU \citep{nair_rectified_2010}, ELU \citep{clevert_fast_2015} and SELU \citep{klambauer_self-normalizing_2017}.
			SELU consistently outperformed ELU and ReLU in terms of training time and network accuracy.
			SELU is computationally more expensive than ReLU but is similar to that of ELU.
			Results in \cref{sec:visual_mesh_results} only include those tested with SELU.

			The network depths used for evaluation were three, five, and nine layers.
			These were chosen as their receptive fields were half, one and two ball radii, respectively.
			This ensured the networks had sufficient contextual information to correctly classify the ball.

			In addition to these Visual Mesh networks, similar CNNs using a regular hexagonal grid were trained.
			These networks allowed a comparison between the Visual Mesh and a network that has equal computational cost due to selecting the same number of pixels.
			This network provides a comparison to an equivalent network without the  constant sample density of the Visual Mesh.

		\subsection{Training}
			The training of these networks was undertaken using the TensorFlow library~\citep{abadi_tensorflow:_2016}.
			The pixel coordinates from the Visual Mesh and the indices of each pixel's six neighbors were used to apply the Visual Mesh at each layer.
			Once this gather step was performed, the neural network steps were undertaken as normal.

			When training these neural networks, the number of ball points and non-ball points were balanced.
			This was achieved by selecting an equal number of points from each class.
			The backpropagation gradients were only calculated from the selected points.

			This method was chosen instead of the traditional method of weighting the gradients intentionally.
			The majority of non-ball points in training images are grass.
			As a result, the initial networks experienced over-fitting on the field.

			Once the initial network was trained, the error in its classification of each point in the image was used as a probability to select that point.
			This resulted in fewer grass points selected in future training.
			This resampling was run twice, with the probabilities added together with a \SI{5}{\percent} baseline probability.
			This greatly improved the accuracy in subsequent training.

			In addition to these networks, five convolutional network architectures were fine-tuned on this dataset.
			These networks were SSD MobileNet and RCNN Inception~V2 trained using TensorFlow \citep{abadi_tensorflow:_2016,huang_speed/accuracy_2016} and YOLOv1 \citep{redmon_you-only_2015}, YOLOv2 \citep{redmon_yolo9000_2017} and YOLOv3 \citep{redmon_yolov3_2018} trained using Darknet\footnote{Darknet \url{http://pjreddie.com/darknet/}}.
			These networks were chosen as they were regarded as some of the fastest real-time networks.

	\subsection{Results}
		\label{sec:visual_mesh_results}

		\subsubsection{Precision}

			\begin{figure}[htbp]
				\centering
				\tikzsetnextfilename{plots}

\pgfplotsset{table/search path={figures/plots/}}

\definecolor{colour1}{HTML}{1b9e77}
\definecolor{colour2}{HTML}{d95f02}
\definecolor{colour3}{HTML}{7570b3}
\definecolor{colour4}{HTML}{e7298a}
\definecolor{colour5}{HTML}{66a61e}
\definecolor{colour6}{HTML}{e6ab02}
\definecolor{colour7}{HTML}{a6761d}
\definecolor{colour8}{HTML}{666666}

\definecolor{distc1}{HTML}{e41a1c}
\definecolor{distc2}{HTML}{377eb8}
\definecolor{distc3}{HTML}{4daf4a}

\addtocounter{figure}{1}

\begin{tikzpicture}[>=latex]
	\begin{axis}[
			name=pr,
			xmin=0.0,
			xmax=1.0,
			ymin=0.0,
			ymax=1.0,
			width=0.5\textwidth,
			xlabel=Recall,
			ylabel=Precision,
			tick label style={/pgf/number format/fixed},
			legend style={at={(1.4,0.5)},anchor=west,draw=none},
			legend cell align=left]

		\addplot[mark=none, color=colour1, solid, thick] table [col sep=comma, x index = {0}, y index = {1}] {data/pr_mesh_3.csv};
		\addlegendentry{Visual Mesh 3L};

		\addplot[mark=none, color=colour1, dashed, thick] table [col sep=comma, x index = {0}, y index = {1}] {data/pr_hex_3.csv};
		\addlegendentry{Hexagonal Mesh 3L};

		\addplot[mark=none, color=colour2, solid, thick] table [col sep=comma, x index = {0}, y index = {1}] {data/pr_mesh_5.csv};
		\addlegendentry{Visual Mesh 5L};

		\addplot[mark=none, color=colour2, dashed, thick] table [col sep=comma, x index = {0}, y index = {1}] {data/pr_hex_5.csv};
		\addlegendentry{Hexagonal Mesh 5L};

		\addplot[mark=none, color=colour3, solid, thick] table [col sep=comma, x index = {0}, y index = {1}] {data/pr_mesh_9.csv};
		\addlegendentry{Visual Mesh 9L};

		\addplot[mark=none, color=colour3, dashed, thick] table [col sep=comma, x index = {0}, y index = {1}] {data/pr_hex_9.csv};
		\addlegendentry{Hexagonal Mesh 9L};

		\addplot[mark=none, color=colour4, dashed, thick] table [col sep=comma, x index = {0}, y index = {1}] {data/pr_yolo.csv};
		\addlegendentry{YOLO};

		\addplot[mark=none, color=colour5, dashed, thick] table [col sep=comma, x index = {0}, y index = {1}] {data/pr_yolov2.csv};
		\addlegendentry{YOLOv2};

		\addplot[mark=none, color=colour6, dashed, thick] table [col sep=comma, x index = {0}, y index = {1}] {data/pr_yolov3.csv};
		\addlegendentry{YOLOv3};

		\addplot[mark=none, color=colour7, dashed, thick] table [col sep=comma, x index = {0}, y index = {1}] {data/pr_ssd_mobilenet.csv};
		\addlegendentry{SSD MobileNet};

		\addplot[mark=none, color=colour8, dashed, thick] table [col sep=comma, x index = {0}, y index = {1}] {data/pr_mask_rcnn.csv};
		\addlegendentry{RCNN Inception~V2};

	\end{axis}

	\node[text width=6cm,align=center,anchor=south] at (pr.north) {\captionof{subfigure}{\label{fig:pr_curve}Precision/Recall}};

	\begin{axis}[
			name=distance,
			width=0.5\textwidth,
			xlabel=Distance (m),
			ylabel=Sample Points,
			at={(0,-0.45\textwidth)},
			tick label style={/pgf/number format/fixed},
			legend entries={$k=5$, $k=4$, $k=3$},
			legend cell align=left]

		\addlegendimage{no markers, solid, black, thick}
		\addlegendimage{no markers, dashed, black, thick}
		\addlegendimage{no markers, dotted, black, thick}

		\addplot[mark=none, color=green, solid, thick] table [col sep=comma, x index = {0}, y index = {1}] {data/dist_5hex.csv};
		\addplot[mark=none, color=green, dashed, thick] table [col sep=comma, x index = {0}, y index = {1}] {data/dist_4hex.csv};
		\addplot[mark=none, color=green, dotted, thick] table [col sep=comma, x index = {0}, y index = {1}] {data/dist_3hex.csv};

		\addplot[mark=none, color=blue, solid, thick] table [col sep=comma, x index = {0}, y index = {1}] {data/dist_5mesh.csv};
		\addplot[mark=none, color=blue, dashed, thick] table [col sep=comma, x index = {0}, y index = {1}] {data/dist_4mesh.csv};
		\addplot[mark=none, color=blue, dotted, thick] table [col sep=comma, x index = {0}, y index = {1}] {data/dist_3mesh.csv};

	\end{axis}

	\begin{axis}[
			name=distance,
			width=0.5\textwidth,
			xlabel=Distance (m),
			ylabel=Sample Points,
			at={(0,-0.45\textwidth)},
			tick label style={/pgf/number format/fixed},
			legend entries={VM, HM},
			legend style={at={(0.23,0.85)},anchor=west},
			legend cell align=left]

			\addlegendimage{no markers, solid, blue, thick};
			\addlegendimage{no markers, solid, green, thick};

			\addplot[mark=none, draw=none, color=green, solid, thick] table [col sep=comma, x index = {0}, y index = {1}] {data/dist_5hex.csv};
			\addplot[mark=none, draw=none, color=green, dashed, thick] table [col sep=comma, x index = {0}, y index = {1}] {data/dist_4hex.csv};
			\addplot[mark=none, draw=none, color=green, dotted, thick] table [col sep=comma, x index = {0}, y index = {1}] {data/dist_3hex.csv};

			\addplot[mark=none, draw=none, color=blue, solid, thick] table [col sep=comma, x index = {0}, y index = {1}] {data/dist_5mesh.csv};
			\addplot[mark=none, draw=none, color=blue, dashed, thick] table [col sep=comma, x index = {0}, y index = {1}] {data/dist_4mesh.csv};
			\addplot[mark=none, draw=none, color=blue, dotted, thick] table [col sep=comma, x index = {0}, y index = {1}] {data/dist_3mesh.csv};
	\end{axis}

	\node[text width=6cm,align=center,anchor=south] at (distance.north) {\captionof{subfigure}{\label{fig:ball_distance_graph}Sample Points/Distance}};

	\begin{axis}[
			name=prmap,
			xmin=0.0,
			xmax=10.0,
			ymin=0.0,
			ymax=1.0,
			at={(0.5\textwidth,-0.45\textwidth)},
			width=0.5\textwidth,
			xlabel=Distance (m),
			ylabel=Mean Average Precision,
			tick label style={/pgf/number format/fixed},
			legend style={at={(1.1,0.5)},anchor=west},
			legend cell align=left
		]

		\addplot[mark=none, color=colour1, solid, thick] table [col sep=comma, x index = {0}, y index = {1}] {data/pr_map_mesh_3.csv};

		\addplot[mark=none, color=colour1, dashed, thick] table [col sep=comma, x index = {0}, y index = {1}] {data/pr_map_hex_3.csv};

		\addplot[mark=none, color=colour2, solid, thick] table [col sep=comma, x index = {0}, y index = {1}] {data/pr_map_mesh_5.csv};

		\addplot[mark=none, color=colour2, dashed, thick] table [col sep=comma, x index = {0}, y index = {1}] {data/pr_map_hex_5.csv};

		\addplot[mark=none, color=colour3, solid, thick] table [col sep=comma, x index = {0}, y index = {1}] {data/pr_map_mesh_9.csv};

		\addplot[mark=none, color=colour3, dashed, thick] table [col sep=comma, x index = {0}, y index = {1}] {data/pr_map_hex_9.csv};

		\addplot[mark=none, color=colour4, dashed, thick] table [col sep=comma, x index = {0}, y index = {1}] {data/pr_map_yolo.csv};

		\addplot[mark=none, color=colour5, dashed, thick] table [col sep=comma, x index = {0}, y index = {1}] {data/pr_map_yolov2.csv};

		\addplot[mark=none, color=colour6, dashed, thick] table [col sep=comma, x index = {0}, y index = {1}] {data/pr_map_yolov3.csv};

		\addplot[mark=none, color=colour7, dashed, thick] table [col sep=comma, x index = {0}, y index = {1}] {data/pr_map_ssd_mobilenet.csv};

		\addplot[mark=none, color=colour8, dashed, thick] table [col sep=comma, x index = {0}, y index = {1}] {data/pr_map_mask_rcnn.csv};
	\end{axis}

	\node[text width=6cm,align=center,anchor=south] at (prmap.north) {\captionof{subfigure}{\label{fig:pr_map}MAP/Distance}};

\end{tikzpicture}

\addtocounter{figure}{-1}
				\caption{
					\label{fig:plots}
					(a) The Precision/Recall curve over all data.\\
					(b) The number of sampled points in an object over distance (VM is Visual Mesh HM is Hexagonal Mesh).\\
					(c) The Average Precision of the detectors over distance.\\
				}
			\end{figure}

			In addition to the Visual and hexagonal meshes, five typical CNNs were also evaluated.
			Their results were measured using a \SI{75}{\percent} IoU.
			\SI{75}{\percent} was chosen as \SI{50}{\percent} was considered a poor match.
			With \SI{50}{\percent} IoU, the center of the detection can be at the edge of the object.

			As shown in \cref{fig:pr_curve}, the accuracy of the Visual Mesh consistently outperforms the hexagonal mesh of an equivalent size.
			Increasing the depth of the network increases its performance.

			The performance of the Visual Mesh remains approximately constant as distance increases.
			However, as shown in \cref{fig:pr_map} the performance of the hexagonal mesh, as well as the other CNNs degrades with increased distance.
			Note that as the generated data was made uniform over pixel size rather than distance, the number of sample images falls off as distance increases.
			The fewer samples increase noise in the plot.

			The performance of the Visual Mesh exceeds the performance of the hexagonal mesh even when the number of points in the object is the same.
			The number of points in both tested networks are equal at \SI{2.5}{\meter}.
			\cref{fig:ball_distance_graph} shows the number of points in the Visual Mesh stays constant over distance, except for a peak at \SI{0}{\meter}.
			This peak is when points are directly below the camera.
			This is a singularity point for the Visual Mesh as it is currently implemented.
			The hexagonal mesh has a decreasing number of points as distance increases.

		\subsubsection{Detections}

\begin{landscape}
	\begin{figure}[htbp]
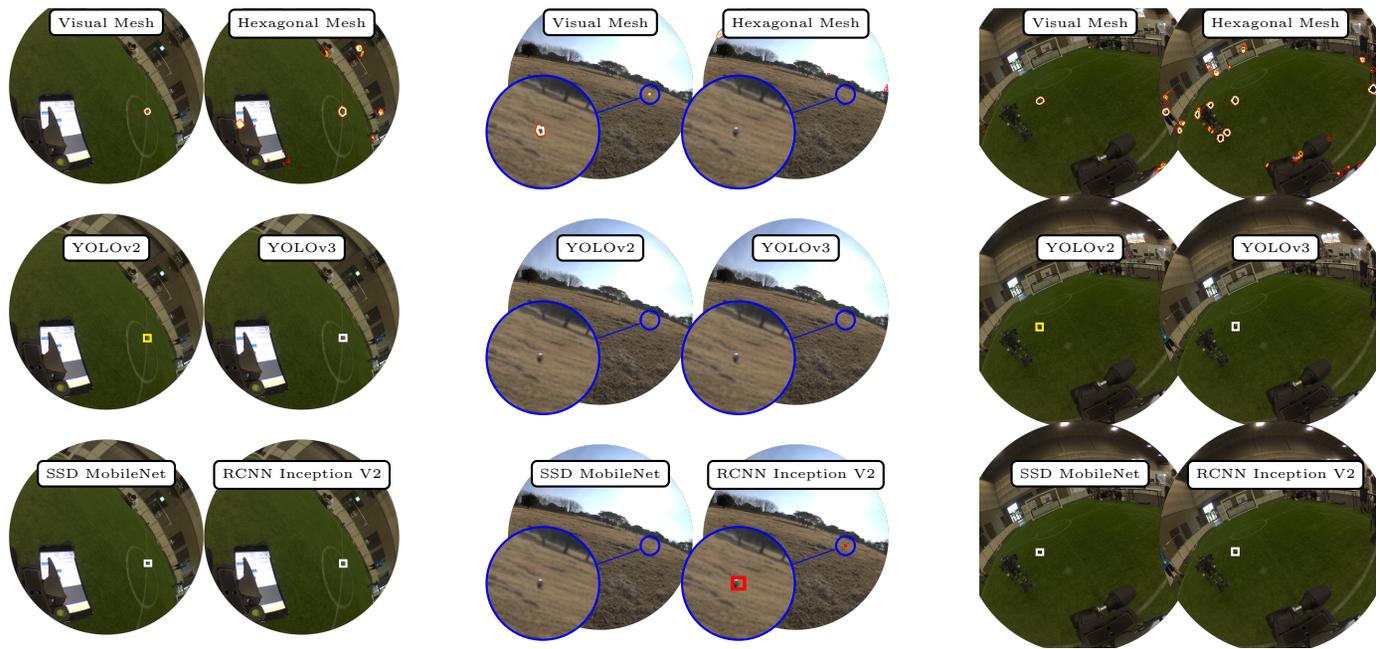

		\centering
		\begin{subfigure}[t]{0.52\textwidth}
			\centering
			\subimport{figures/detections/}{example_detections_1}

			\caption{
				\label{fig:ball_detection_images_1}
				Detection on an empty field.
				All networks achieve acceptable performance.
				However, the hexagonal mesh has some false positives.
			}
		\end{subfigure}
		\hfill
		\begin{subfigure}[t]{0.52\textwidth}
			\centering
			\subimport{figures/detections/}{example_detections_2}

			\caption{
				\label{fig:ball_detection_images_2}
				Detections at extreme distance.
				Only the Visual Mesh achieves a good detection.
				RCNN Inception~V2 detects but IoU of the bounding box is $<50\%$.
			}
		\end{subfigure}
		\hfill
		\begin{subfigure}[t]{0.52\textwidth}
			\centering
			\subimport{figures/detections/}{example_detections_3}

			\caption{
				\label{fig:ball_detection_images_3}
				Detection on a field with a robot.
				The hexagonal mesh has many false positives on the robot, while the other detectors perform well.
			}
		\end{subfigure}

		\caption{
			\label{fig:ball_detection_images}
			Example detections where confidence intervals are represented by colors: Red $>50\%$, Yellow $>75\%$ and White $>90\%$
		}
	\end{figure}
\end{landscape}

			\cref{fig:ball_detection_images_1} shows a typical set of detections from each of the trained networks.
			YOLOv1 is omitted as it performs strictly worse than YOLOv2.
			The Visual Mesh has a good detection while the hexagonal mesh has several false positives.
			The five other networks all detect the ball.
			YOLOv2 has a lower confidence than the other networks on the dataset.
			SSD MobileNet, YOLOv2 and YOLOv3's bounding boxes are less accurate across the dataset.

			The Visual Mesh excels at distant detections as shown in \cref{fig:ball_detection_images_2}.
			Except for the Visual Mesh and RCNN Inception~V2, none of the other networks detect the ball.
			RCNN Inception~V2 has a poorly fitted bounding box.
			This is typical of distant balls in the dataset.

			\cref{fig:ball_detection_images_3} shows how the Visual Mesh is able to use scale to identify target objects.
			The hexagonal mesh found many false positives on objects that had a different size than expected, but similar appearance as the target.

		\subsubsection{Execution performance}

			Each network was tested on the CPU and GPU from the Intel NUC7i7BNH as well as on an NVIDIA 1080Ti.
			The input images were $1280\times 1024$ for all networks.
			SSD MobileNet and RCNN Inception~V2 were not executed on the Intel GPU as TensorFlow does not support OpenCL at this time.
			YOLOv3 was not executed as it is not supported by the OpenCL version of Darknet.
			\cref{tab:visualmesh_performance} summarizes the results with respect to execution time.
			The times for all networks are measured from when the image is first sent to the algorithm until the inferences are returned.
			Therefore, the time taken to project Visual Mesh points is included.

			\begin{table}[htbp]
				\centering

				\caption{
					\label{tab:visualmesh_performance}
					Execution performance: For the Visual Mesh on the Iris Plus Graphics and the NVIDIA 1080Ti the device utilization was \SI{70}{\percent} and \SI{35}{\percent} respectively.
					For all other cases utilization was at \SI{100}{\percent}.
				}

				\begin{tabular}{l|S[table-format=5.5,table-align-text-post=false]|S[table-format=3.2,table-align-text-post=false]|S[table-format=2.4,table-align-text-post=false]}
					\toprule
					                  & {Intel Core}               & {Intel Iris Plus}                & {NVIDIA}           \\
					                  & {i7 7567U}                 & {Graphics 650}                   & {1080Ti}           \\ \midrule
					Visual Mesh 5     &    1.64 \si{\milli\second} &   2.10 \si{\milli\second}  &  2.18 \si{\milli\second} \\
					Visual Mesh 9     &    2.44 \si{\milli\second} &   2.48 \si{\milli\second}  &  2.25 \si{\milli\second} \\
					YOLOv1            & 1468.24 \si{\milli\second} & 721.13 \si{\milli\second}  & 17.55 \si{\milli\second} \\
					YOLOv2            & 1221.49 \si{\milli\second} & 613.73 \si{\milli\second}  & 16.13 \si{\milli\second} \\
					YOLOv3            & 2651.33 \si{\milli\second} & {N/A}                      & 19.00 \si{\milli\second} \\
					SSD MobileNet     &   37.76 \si{\milli\second} & {N/A}                      & 11.32 \si{\milli\second} \\
					RCNN Inception~V2 & 1521.32 \si{\milli\second} & {N/A}                      & 47.75 \si{\milli\second} \\ \bottomrule
				\end{tabular}
			\end{table}

	\subsection{Discussion and Conclusion}

		The results for the Visual Mesh show that consistent feature density improves the accuracy of the network.
		When the Visual Mesh and the hexagonal mesh had an equal number of points on the ball the Visual Mesh was more accurate.
		As distance increased, the accuracy of the hexagonal mesh degraded while the Visual Mesh remained consistent.
		This degradation can also be seen in other networks as accuracy declines over distance.

		The nine-layer Visual Mesh is used for the following comparisons.
		It provided the highest accuracy and its computational performance was not significantly worse than the five-layer Visual Mesh.

		RCNN Inception~V2 and YOLOv3 performed the best of the other networks tested.
		While the other CNNs failed to detect distant objects, these networks continued to detect them.
		However, the bounding boxes became increasingly inaccurate.
		At a lower IoU threshold they have a higher detection rate.
		Visual Mesh exceeds their performance after \SI{4}{\meter}.

		As seen in \cref{tab:visualmesh_performance}, the execution performance of the Visual Mesh exceeds that of the other convolutional networks.
		Of these networks, only SSD MobileNet and the Visual Mesh could be considered for real-time use on resource-constrained systems.
		The performance of the Visual Mesh is fast enough that the transfer times of images is a significant factor for GPU based computation.
		The NUC's CPU outperforms its GPU for the five-layer Visual Mesh because of this.
		The NVIDIA 1080Ti also suffers this effect, resulting in only \SI{35}{\percent} utilization.

		The Visual Mesh has a number of advantages beyond its accuracy and speed.
		As objects always have a similar number of points additional post-processing options are available to improve accuracy.
		Within detected areas, metrics such as graph diameter can be used to filter out irrelevant areas.
		Additionally the best fitting subgraph can be used to remove invalid points in a detection.

		Higher resolution does not increase the computational cost of the Visual Mesh.
		The number of points that are projected onto the image does not change for the same camera lens and orientation.
		However, increased resolution will allow the Visual Mesh to project points that are a greater distance from the camera.
		If the resolution of the camera is insufficient for the level of detail requested, the Visual Mesh will begin sampling the same pixel multiple times.
		The Visual Mesh is still accurate with limited amounts of this duplicated data.
		However, as the amount of information decreases, the accuracy of the network will decline.

		As the distance to objects increases, typical networks must learn to account for the differences in scale that occur.
		Often these differences are not well represented in the training data or, can be biased in the training data.
		This can require additional training data to be generated by scaling.
		For the Visual Mesh, this is not necessary as the object will always appear the same size.
		This reduces the complexity of training as well as the complexity of the required network.

		As the Visual Mesh is always oriented relative to the observation plane, the resulting network is better able to handle changes in the orientation of the camera.
		This form of transformational invariance only applies to rotations in the camera, not rotations on the object.
		This can reduce the amount of training required if the object can be assumed in a particular rotation.
		For example, if extended to detect the goal posts the training data would not need to be modified for different orientations of the camera as they would always be normal to the observation plane.
		Without this invariance, training for goal posts would have to include multiple orientations.

		Networks based on the Visual Mesh also have an independence to the lens used.
		As the sample points are always in the same place in the world, changing to a different lens geometry does not change the points.
		This makes it easier to train using data from different lenses and apply trained networks to new lenses.

		The presented formulation of the Visual Mesh has two primary limitations.
		One concern is that it cannot function when the height of objects are greater than or equal to the height of the camera.
		In these cases, the Visual Mesh correctly predicts that all objects are visible on the horizon.
		This results in a single line of points.
		In practice, this does not afford good detection performance.
		The second limitation is that objects that are directly below the camera fall into a singularity.
		When in this singularity, twice as many points intersect with the objects until they move beyond this position.
		This increases the complexity that the Visual Mesh must learn.

		The training and execution code for the Visual Mesh is available at \url{https://github.com/Fastcode/VisualMesh}.

\subsection*{Acknowledgments}
	TH was supported by a Australian Government Research Training Program scholarship and a completion scholarship from 4Tel Pty. Ltd.

\bibliographystyle{splncsnat}
\bibliography{references}

\begin{thebibliography}{13}
\providecommand{\natexlab}[1]{#1}
\providecommand{\url}[1]{\texttt{#1}}
\providecommand{\urlprefix}{}

\bibitem[{Abadi and {et al.}(2016)}]{abadi_tensorflow:_2016}
Abadi, M., {et al.}: Tensorflow: A system for large-scale machine learning.
\newblock In: Proceedings of the 12th USENIX Conference on Operating Systems
  Design and Implementation. pp. 265--283. {OSDI}'16, USENIX Association,
  Berkeley, CA, USA (2016)

\bibitem[{Albani et~al.(2017)Albani, Youssef, Suriani, Nardi, and
  Bloisi}]{albani_a-deep_2017}
Albani, D., Youssef, A., Suriani, V., Nardi, D., Bloisi, D.D.: A deep learning
  approach for object recognition with {NAO} soccer robots.
\newblock In: Behnke, S., Sheh, R., Sarıel, S., Lee, D.D. (eds.) RoboCup 2016:
  Robot World Cup XX. pp. 392--403. Springer International Publishing, Cham
  (2017)

\bibitem[{Bloisi et~al.(2017)Bloisi, Duchetto, Manoni, and
  Suriani}]{bloisi_machine_2017}
Bloisi, D., Duchetto, F.D., Manoni, T., Suriani, V.: Machine learning for
  realistic ball detection in robocup {SPL}.
\newblock CoRR abs/1707.03628 (2017)

\bibitem[{Clevert et~al.(2015)Clevert, Unterthiner, and
  Hochreiter}]{clevert_fast_2015}
Clevert, D.A., Unterthiner, T., Hochreiter, S.: Fast and accurate deep network
  learning by exponential linear units (elus).
\newblock CoRR abs/1511.07289 (2015)

\bibitem[{Cruz et~al.(2017)Cruz, Lobos-Tsunekawa, and Ruiz-del
  Solar}]{cruz_using_2017}
Cruz, N., Lobos-Tsunekawa, K., Ruiz-del Solar, J.: Using convolutional neural
  networks in robots with limited computational resources: Detecting {NAO}
  robots while playing soccer.
\newblock CoRR abs/1706.06702 (2017)

\bibitem[{Huang et~al.(2016)Huang, Rathod, Sun, Zhu, Korattikara, Fathi,
  Fischer, Wojna, Song, Guadarrama, and Murphy}]{huang_speed/accuracy_2016}
Huang, J., Rathod, V., Sun, C., Zhu, M., Korattikara, A., Fathi, A., Fischer,
  I., Wojna, Z., Song, Y., Guadarrama, S., Murphy, K.: Speed/accuracy
  trade-offs for modern convolutional object detectors.
\newblock CoRR abs/1611.10012 (2016)

\bibitem[{Javadi et~al.(2018)Javadi, Azar, Azami, Shiry, Ghidary, and
  Baltes}]{javadi_humanoid_2017}
Javadi, M., Azar, S.M., Azami, S., Shiry, S., Ghidary, S.S., Baltes, J.:
  Humanoid robot detection using deep learning: A speed-accuracy tradeoff.
\newblock In: RoboCup 2017: Robot World Cup XXI. Springer International
  Publishing (2018), in press

\bibitem[{Klambauer et~al.(2017)Klambauer, Unterthiner, Mayr, and
  Hochreiter}]{klambauer_self-normalizing_2017}
Klambauer, G., Unterthiner, T., Mayr, A., Hochreiter, S.: Self-normalizing
  neural networks.
\newblock In: Guyon, I., Luxburg, U.V., Bengio, S., Wallach, H., Fergus, R.,
  Vishwanathan, S., Garnett, R. (eds.) Advances in Neural Information
  Processing Systems 30, pp. 971--980. Curran Associates, Inc. (2017)

\bibitem[{Nair and Hinton(2010)}]{nair_rectified_2010}
Nair, V., Hinton, G.E.: Rectified linear units improve restricted boltzmann
  machines.
\newblock In: Proceedings of the 27th International Conference on International
  Conference on Machine Learning. pp. 807--814. {ICML}'10, Omnipress, USA
  (2010)

\bibitem[{Redmon et~al.(2015)Redmon, Divvala, Girshick, and
  Farhadi}]{redmon_you-only_2015}
Redmon, J., Divvala, S.K., Girshick, R.B., Farhadi, A.: You only look once:
  Unified, real-time object detection.
\newblock CoRR abs/1506.02640 (2015)

\bibitem[{Redmon and Farhadi(2017)}]{redmon_yolo9000_2017}
Redmon, J., Farhadi, A.: {YOLO}9000: Better, faster, stronger.
\newblock In: 2017 IEEE Conference on Computer Vision and Pattern Recognition
  (CVPR). pp. 6517--6525 (2017)

\bibitem[{Redmon and Farhadi(2018)}]{redmon_yolov3_2018}
Redmon, J., Farhadi, A.: {YOLO}v3: An incremental improvement.
\newblock CoRR abs/1804.02767 (2018)

\bibitem[{Speck et~al.(2017)Speck, Barros, Weber, and
  Wermter}]{speck_ball_2016}
Speck, D., Barros, P., Weber, C., Wermter, S.: Ball localization for robocup
  soccer using convolutional neural networks.
\newblock In: Behnke, S., Sheh, R., Sarıel, S., Lee, D.D. (eds.) RoboCup 2016:
  Robot World Cup XX. pp. 19--30. Springer International Publishing, Cham
  (2017)

\end{thebibliography}

\end{document}